# GENDER IDENTITY AND LEXICAL VARIATION IN SOCIAL MEDIA[1]

**David Bamman, Jacob Eisenstein, and Tyler Schnoebelen**

**Abstract**: We present a study of the relationship between gender, linguistic style, and social networks, using a novel corpus of 14,000 Twitter users. Prior quantitative work on gender often treats this social variable as a female/male binary; we argue for a more nuanced approach. By clustering Twitter users, we find a natural decomposition of the dataset into various styles and topical interests. Many clusters have strong gender orientations, but their use of linguistic resources sometimes directly conflicts with the population-level language statistics. We view these clusters as a more accurate reflection of the multifaceted nature of gendered language styles. Previous corpus-based work has also had little to say about individuals whose linguistic styles defy population-level gender patterns. To identify such individuals, we train a statistical classifier, and measure the classifier confidence for each individual in the dataset. Examining individuals whose language does not match the classifier's model for their gender, we find that they have social networks that include significantly fewer same-gender social connections, and that in general, social network homophily is correlated with the use of same-gender language markers. Pairing computational methods and social theory thus offers a new perspective on how gender emerges as individuals position themselves relative to audiences, topics, and mainstream gender norms.
[206 words]

# GENDER IDENTITY AND LEXICAL VARIATION IN SOCIAL MEDIA[i]

**David Bamman, Jacob Eisenstein and Tyler Schnoebelen**

The increasing prevalence of online social media for informal communication has enabled large-scale statistical analysis of the connection between language and social variables such as gender, age, race, and geographical origin. Whether the goal of such research is to understand stylistic differences or to build predictive models of 'latent attributes', there is often an implicit assumption that linguistic choices are associated with immutable and essential categories of people. Indeed, strong aggregate correlations between language and such categories enable predictive models that are disarmingly accurate. But this gives an oversimplified and misleading picture of how language conveys personal identity.

In this paper, we present a study of the relationship between gender, language, and social network connections in social media text. We use a novel corpus of more than 14,000 individuals on the microblog site Twitter, and perform a computational analysis of the impact of gender on both their lexical choices and their social networks. We address two limitations of previous quantitative analyses of language and gender.

First, previous work has focused on words that distinguish women and men solely by gender. This disregards theoretical arguments and qualitative evidence that gender can be enacted through a diversity of styles and stances. Using clustering, we identify a range of linguistic styles and topical interests among the authors in our dataset. Many of these clusters have strong gender orientations — demonstrating the strong relationship between language and

gender — but the clusters reveal many possible alignments between linguistic resources and gender, sometimes directly conflicting with the aggregate statistics for the gender that dominates the cluster.

Second, previous quantitative work has had little to say about individuals whose word usage defies aggregate language-gender statistics. We identify such individuals by training an accurate statistical classifier to predict gender based on language, and then focusing on people for whom these predictions fail. We find a significant correlation between the use of mainstream gendered language — as represented by classifier confidence — and the extent to which an individual's social network is made up of same-gender individuals: people whose gender is classified incorrectly have social networks that are much less homophilous. Social network gender homophily and the use of mainstream gendered linguistic features are closely linked, even after controlling for author gender.

## BACKGROUND

Gender is a pervasive topic in the history of sociolinguistics. The concepts of 'standard' and 'vernacular' have been repeatedly recruited to characterize gender differences in language (Cheshire 2002; Coates and Cameron 1989; Eckert and McConnell-Ginet 1999). Over time, differences in the use of conventional and innovative features have come to be seen as part of acquiring and deploying symbolic capital (see also Holmes 1997). As these lines of thinking have developed, many sociolinguists have come to see gender as constructed, maintained, and disrupted by linguistic practices, which, in turn, shape language. However, this theoretical outlook has been applied mainly in small-scale qualitative studies, and is absent from much of the work on larger corpora. Such datasets, typically derived from social media, make it possible to analyze the frequency of individual words. This has led to a wave of computational research on the automatic identification of 'latent attributes' such as gender, age, and regional origin (e.g., Rao, Yarowsky, Shreevats, and Gupta 2010). Coming from the computer science research tradition, much of this work is built around an instrumentalist paradigm that emphasizes prediction of latent attributes from text. In this methodology, accurate predictions justify a post hoc analysis to identify the words that are the most effective predictors; these words are then assembled into groups. This reverses the direction of earlier corpus-based work in which word classes are defined in advance, and then compared quantitatively across genders.

In one such study, Argamon, Koppel, Pennebaker, and Schler (2007) assembled 19,320 English blogs (681,288 posts, 140 million words), and built a predictive model of gender that achieves 80.5% accuracy. A post-hoc factor analysis found that content-related factors are used more often by men, while style-related factors are used more by women. More recent studies have focused on Twitter, a microblogging platform. Rao et al. (2010) trained a classifier on a dataset of posts ('tweets') by 1,000 authors. They found that women used more emoticons, ellipses (...), expressive lengthening (*nooo waaay*), complex punctuation (*!!* and *?!*), and transcriptions of backchannels (*ah*, *hmm*). The only words strongly associated with men were affirmations like *yeah* and *yea*. In this study, the author pool was built from individuals with metadata connections to explicitly gendered entities: sororities, fraternities, and hygiene products. Assumptions about gender were thus built directly into the data acquisition methodology, which is destined to focus on individuals with very specific types of gendered identities. Broadly similar results were found by Burger, Henderson, Kim, and Zarrella et al. (2011), who identified author gender by linking 184,000 Twitter accounts to blog profiles with gender metadata. Remarkably, they found that automatic prediction of author gender is more accurate that the judgments of human raters.

Contemporary computational literature often distinguishes men and women on pragmatic dimensions of ‘informativeness’ and ‘involvement’ (Argamon, Koppel, Fine, and Shimoni 2003), based on earlier corpus-based contrasts of written and spoken genres (Biber 1995; Chafe 1982). The involvement dimension consists of linguistic resources that create interactions between speakers and their audiences; the informational dimension consists of resources that communicate propositional content. Early work compared frequencies of large word classes, such as parts-of-speech: the involvement dimension includes first and second person pronouns, present tense verbs, and contractions (Biber 1988, 1995; Tannen 1982), while the informational dimension includes prepositions, attributive adjectives, and longer words. Informational word classes were found to be used preferentially by men, while involvement and interaction are associated with women (Argamon et al. 2003; Herring and Paolillo 2006; Schler, Koppel, Argamon, and Pennebaker 2006). A related distinction is contextuality: males are seen as preferring a ‘formal’ and ‘explicit’ style, while females are seen as preferring a style that is more ‘deictic’ and ‘contextual’ (Heylighen and Dewaele 2002; Nowson, Oberlander, and Gill 2005).

Herring and Paolillo (2006) offer reason for caution in extending predictive results to descriptive statements about the linguistic resources preferred by women and men. Using blog data, they replicated the association of informational and involvement word classes with male and female authors. But after controlling for the genre of the blog, the gender differences disappeared. Women were more likely to write 'diary' blogs, while men were more likely to write 'filter' blogs, linking to external content. The involvement and informational word classes were associated with these genres, and the genres were in turn associated with gender. But within each genre, there were no significant gender differences in the frequency of the word classes. Had Herring and Paolillo simply aggregated all blog posts without regard to genre, they would have missed this key mediating factor.

Quantitative analysis of language and gender requires other, more subtle forms of aggregation — not least, the description of individuals as women and men. While these social categories are real, choosing this binary opposition as a starting point constrains the set of possible conclusions. It is illustrative to consider how gender interacts with other aspects of personal identity. Eckert and McConnell-Ginet (1995) examined the interaction between gender and the local categories of school-oriented 'jocks' and anti-school 'burnouts', finding that boys were less standard than girls in general, but that the most non-standard language was employed by a group of 'burned-out burnout' girls.

The complex role of gender in larger configurations of personal identity poses problems for quantitative analyses that aggregate individuals by gender alone. Eckert (2008) and others have argued that the social meaning of linguistic variables depends crucially on the social and linguistic context in which they are deployed. Rather than describing variables like ING/IN as a direct reflection of gender or class, they can be seen as reflecting a field of meanings: educated/uneducated, effortful/easygoing, articulate/inarticulate, pretentious/unpretentious, formal/relaxed, and so on. The indexical field of a linguistic resource is used to create various stances and personae, which are connected both to global categories like race and gender, but also to local distinctions like jocks versus burnouts. This view has roots in Butler's (1990:179) casting of gender as a stylized repetition of acts, creating a relationship between (at least) an individual, an audience, and a topic (Schnoebelen 2012).

This theoretical perspective leads to anti-essentialist conclusions: gender and other social categories are performances, and these categories are performed differently in different situations (see also Coates 1996; Hall 1995). Empirical work in this tradition can shed light on the ways in which the interaction between language and gender are mediated by situational contexts. For example, Barrett (1999) described how African American drag queens appropriate 'white woman' speech, while rapidly shifting between linguistic styles. Goodwin (1990) examined how boys and girls employ sharply differentiated styles in some activities, and more homogeneous styles in others. Kiesling (2004) argued that the term *dude* allows men to meet needs for 'homosocial solidarity' and closeness, without challenging their heterosexuality. Each of these studies demonstrates a richness of interactions between language, gender, and situational context.

Studies that focus on the social construction of gender in specific conversational contexts are necessarily qualitative, but our goal is to bring the spirit of this work to large-scale quantitative analysis. We see these methodologies as complementary. Qualitative analysis can point to phenomena that can be quantitatively pursued at much larger scale. Conversely, exploratory quantitative analysis can identify candidates for closer qualitative reading into the depth and subtlety of social meaning in context.

## DATA

Our research is supported by a dataset of microblog posts from the social media service Twitter, which allows users to post 140-character messages. Each message appears in the newsfeeds of individuals who have chosen to follow the author, though by default the messages are publicly available to anyone on the Internet. We choose Twitter for several reasons. Unlike Facebook, the majority of content on Twitter is explicitly public. Unlike blogs, Twitter data is encoded in a single format, facilitating large-scale data collection. Twitter has relatively broad penetration across different ethnicities, genders, and income levels. The Pew Research Center has repeatedly polled the demographics of Twitter (Smith and Brewer 2012), finding: nearly identical usage among women (15% of female internet users are on Twitter) and men (14%); high usage among non-Hispanic Blacks (28%); an even distribution across income and education levels; higher usage among young adults (26% for ages 18-29, 4% for ages 65+).

Large numbers of messages may be collected using Twitter's streaming API, which delivers a sample from the complete stream of public messages. We used this API to gather a corpus from Twitter over a period of six months, between January and June, 2011. Our goal was to collect text that is representative of American English, so we included only messages from authors located in the United States. Full-time non-English users were filtered out by requiring that all authors in the corpus use at least 50 of the 1,000 most common words in the US sample overall (predominantly English terms).

We further filtered our sample to only those individuals who are actively engaging with their social network. Twitter contains an explicit social network in the links between individuals who have chosen to receive each other's messages. However, Kwak, Lee, Park, and Moon (2010) found that only 22 percent of such links are reciprocal, and that a small number of hubs account for a high proportion of the total number of links. Instead, we define a social network based on direct, mutual interactions. In Twitter, it is possible to address a public message towards another user by prepending the @ symbol before the recipient's user name. We build an undirected network of these links. To ensure that the network is mutual and as close of a proxy to a real social network as possible, we form a link between two users only if we observe at least two mentions (one in each direction) separated by at least two weeks. This filters spam accounts, unrequited mentions (e.g., users attempting to attract the attention of celebrities), and one-time conversations. We selected only those users with between four and 100 mutual-mention friends. The upper bound helps avoid 'broadcast-oriented' Twitter accounts such as news media, corporations, and celebrities.

To assign gender to user accounts, we first estimated the distribution of gender over individual names using historical census information from the US Social Security Administration, taking the gender of a first name to be its majority count in the data. We only select users with first names that occur over 1,000 times in the census data (approximately 9,000 names), the most infrequent of which include *Cherylann*, *Kailin* and *Zeno*. After applying all these filters, the resulting dataset contains 14,464 users and 9,212,118 tweets.

Some names are ambiguous by gender, but in our dataset, such ambiguity is rare: the median user has a name that is 99.6% associated with its majority gender; 95% of all users have name that is at least 85% associated with its majority gender. We assume that users tend to self-report their true name; while this may be largely true on aggregate, there are bound to be exceptions. Our analysis therefore focuses on aggregate trends and not individual case studies. A second potential concern is that social categories are not equally relevant in every utterance. But while this is certainly true in some cases, it is not true on aggregate — otherwise, accurate gender prediction from text would not be possible. Later, we address this issue by analyzing the social behavior of individuals whose language is not easily associated with their gender.

## LEXICAL MARKERS OF GENDER

We begin with an analysis of the lexical markers of gender in our new microblog dataset, taking the standard computational approach of aggregating authors into male and female genders. We build a predictive model based on bag-of-words features, and then identify the most salient lexical markers of each gender. The purpose is to replicate prior work, and to set the stage for the remainder of the paper, which probes the blind spots of this sort of analysis.

To quantify the strength of the relationship between gender and language in our data, we train a logistic regression classifier. Logistic regression is the statistical technique at the core of variable rule analysis (Tagliamonte 2006).

However, our application here is somewhat different from the traditional variationist approach: the dependent variable is the author gender; the independent variables are the 10,000 most frequent lexical items in the corpus, which include individual words and word-like items such as emoticons and punctuation.[2] In traditional variationist applications of logistic regression, gender would be an independent variable and it would be used to predict some linguistic variable. But we reverse this setup because the relevant linguistic variables are not known a priori; this method allows us to discover them from text. Accommodating such a large number of independent variables involves estimating a large number of parameters (one for each word type in the vocabulary). One risk in such high-dimensional settings is **overfitting** -- learning parameter values that perfectly describe the training data, but failing to generalize to new data. In our predictive setting, this may occur, for example, if a single word type (like "John") is used three times by men and zero times by women; an overfit model would have very high confidence that an individual who uses this term is a man, regardless of the other words that they use. To avoid this, we adopt the standard machine learning technique of **regularization**, which dampens the effect of any individual variable (Hastie, Tibshirani, and Friedman, 2009). A single regularization parameter controls the tradeoff between perfectly describing the training data and generalizing to unseen data; we tune this parameter on a held-out development set.

In order to evaluate the accuracy of our classifier on new data that it has not seen before, we conduct a ten-fold cross-validation in which we randomly divide the full dataset into ten parts, train our model on 80% of that data (eight parts), tune the regularization parameter on 10% (one part), and then predict the gender of the remaining 10% (one part). We conduct this test ten times on different partitions of the data, and calculate the overall accuracy as the average of these ten tests. The accuracy in gender prediction by this method is 88.0%, which is state of the art compared with gender prediction on similar datasets (Burger et al. 2011). While more expressive features might perform better still, the high accuracy of lexical features shows that they capture a great deal of language's predictive power with regard to gender.

[2] All words were converted to lower-case but no other preprocessing or stopword filtering was performed.

Our next analysis identifies the words most strongly associated with each gender. We calculate this with a statistical hypothesis test; for each word, we count the fraction of men and women who use the term, and compare those two ratios with the fraction of all people who use it (which corresponds to a null hypothesis in which the word's frequency does not depend on gender). The words for which the ratio of men or women is most dissimilar to the overall ratio -- i.e., for which null hypothesis is a particularly improbable explanation -- are identified as those most strongly gendered; more details are found in Appendix I. Because we are computing statistical hypothesis tests for thousands of different events, we apply the Bonferroni correction for multiple comparisons (Dunn 1961). Even with the correction, more than 500 terms are significantly associated with each gender; we limit our consideration to the 500 terms for each gender with the lowest p-values.

[Insert Table 1 about here]

Table 1 compares our class-level findings with previous results. We note a few specific details about some of the classes. **Pronouns** are generally associated with female authors, including alternative spellings *u, ur*, *yr.* All of the **emotion terms** (*sad, love, glad*, etc.) and **emoticons** that appear as gender markers are associated with female authors, including some that the prior literature found to be neutral or male: *:) :D* and *;)*. Of the **kinship terms** that are gender markers, most are associated with female authors (*mom*, *mommy*, *sister*, *daughter*, *aunt*, *auntie*, *grandma*, *kids*, *child*, *dad*, *husband*, *hubs,* etc.). Only a few kinship-related terms are associated with male authors — *wife*, *wife's*, *bro*, *bruh*, *bros*, and *brotha* — though many of these may be better described as friendship terms, with corresponding female markers *bestie*, *bff*, and *bffs* (*best friends forever*). Several **abbreviations** like *lol* and *omg* appear as female markers, as do ellipses, expressive lengthening (e.g., *coooooool*), exclamation marks, question marks, and backchannel sounds like *ah*, *hmmm*, *ugh*, and *grr*. Hesitation words *um* and *umm* are also associated with female authors, replicating the analysis of speed dating speech by Acton (2011). The **assent terms** *okay*, *yes*, *yess*, *yesss*, *yessss* are all female markers, though *yessir* is a male marker. **Negation terms** *nooo*, *noooo*, and *cannot* are female markers, while *nah*, *nobody*, and *ain't* are male markers. **Swears and taboo words** are more often associated with male authors; the anti-swear *darn* is a female marker. This gendered distinction between mild and strong swear words was previously reported by McEnery (2005). Our analysis did not show strong gender associations for standard **prepositions,** but a few alternative spellings had strong gender associations: *2* (a male marker) is often used as a homophone for *to*; an abbreviated form of *with* appears in the female markers *w/a*, *w/the*, *w/my*. The only **conjunction** that displays significant gender association is *&*, associated with female authors. No **articles or determiners** are found to be significant markers.

These findings are generally in concert with previous research. Yet any systematization of these word-level gender differences into dimensions of standardness or expressiveness faces difficulties. The argument that female language is more expressive is supported by lengthenings like *yesss* and *nooo*, but swear words should also be seen as expressive, and they are generally preferred by men. The rejection of swear words by female authors may seem to indicate a greater tendency towards standard or prestige language, but this is contradicted by abbreviations like *omg* and *lol*.

The word classes defined in prior work failed to capture some of the most salient phenomena in our data, such as the tendency for proper nouns to be more often used by men (*apple's*, *iphone*, *lebron*) and for non-standard spelling to be used more frequently by women (*vacay*, *yayyy*, *lol*). We developed an alternative categorization, with the criterion that each word be unambiguously classifiable into a single category. After identifying eight categories, two of the paper's authors individually categorized each of the 10,000 most frequent terms in the corpus. The initial agreement was 90.0%; disagreements were resolved by discussion between all three authors. The categories are: **named entities** (e.g., *apple's, nba, steve*), including abbreviations such as *fb* (Facebook); **taboo words**; **numbers** (e.g., *2010*, *3-0*, *500*); **hashtags**, a Twitter convention of prepending # to make a searchable keyword; **punctuation**, including individual punctuation marks but not include emoticons or multi-character strings like *!!!;* **dictionary** words found in a standard dictionary and not listed as 'slang', 'vulgar', as proper nouns, or as acronyms; **pronounceable non-dictionary words** (e.g., *nah*, *haha*, *lol),* including contractions written without apostrophes (e.g. *dont*, *cant*); **non-pronounceable non-dictionary words,** which must be spelled out or described to be used in speech, including emoticons and abbreviations (e.g., *omg*, *;)*, *api*).

The ordered list constitutes a pipeline, organized by priority. Each word is placed in the first category it matches (with the following order: named entities > taboo words > numbers > hashtags > punctuation > dictionary > pronounceable non-dictionary > non-pronounceable non-dictionary). For example, although *#fb* is a hashtag and must be spelled out to be pronounced, it is treated as a named entity because that category is most salient. Homographs across several categories were judged by examining a set of random tweets, and the most frequent sense was used to determine the categorization: while *idol* is a dictionary word, it is coded as a named entity because a majority of uses refer to the television program American Idol.

[Insert Table 2 about here]

Table 2 shows the frequency of each category by gender. Due to the size of the dataset, all differences are statistically significant at $p < 0.01$, but many of these differences would be difficult to notice without quantitative analysis. Men mention named entities about 30% more often than women do, and women use emoticons and abbreviations 40% more often than men do. The contrast of named entities versus emoticons may seem to offer evidence for proposed high-level distinctions such as information versus involvement. However, we urge caution. The 'involvement' dimension is characterized by the engagement between the writer/speaker and the audience, which is why involvement is often measured by first and second person pronoun frequency (e.g. Biber 1988). Named entities describe concrete referents, and thus may be thought of as informational, rather than involved; on this view, they are not used to reveal the self or to engage with others. But many of the named entities in our list refer to sports figures and teams, and are thus key components of identity and engagement for their fans (Meân 2001). While it is undeniable that many words have strong statistical associations with gender, the direct association of word types with high-level dimensions remains problematic.

Argamon et al. (2003: 332) note that 'it is notoriously difficult to unambiguously map given linguistic markers to communicative function; we use the terms 'involved' and 'informational' ... simply as a suggestive label for a correlated set of lexical features.' This is an important caveat, and the utility of discussing — and naming — groups of correlated lexical items is undeniable. Nonetheless, it is difficult to be satisfied with an analysis that permits abstract discourse categories like 'involvement/information' and abstract identity categories like 'male/female' to get so tightly coupled that we are left with 'women are involved, men are informational.' In the next section, we build on this notion of correlated lexical features, finding clusters of authors who tend to use similar words. We evaluate the relationship between gender and author clusters, and reconsider language-gender associations in a model that permits gender to be enacted in multiple ways.

## CLUSTERS OF AUTHORS

Automatic text classification makes no assumptions about how or why linguistic resources become predictive of each gender; it simply demonstrates a lower bound on the predictive power that those resources contain. It is the post hoc analysis — identifying lists of words that are most strongly associated with each gender — that smuggles in an additional assumption of direct alignment between linguistic resources and gender. A broad literature of theoretical and empirical work argues that the relationship between language and gender can only be accurately characterized in terms of *situated* meanings, which construct gender through a variety of stances, styles, and personae (Eckert 2008; McConnell-Ginet 2011; Ochs 1992; Schiffrin 1996).

Is it possible to use quantitative methods without positing a direct mapping between gender and linguistic resources? In this section, we revisit the lexical analysis with more delicate tools. Setting gender aside, we use clustering to identify groups of authors who use similar sets of words. In principle such clusters might be completely orthogonal to gender; for example, they might simply correspond to broad areas of interest. But most of the clusters display a strong gender orientation, demonstrating the multiple ways of enacting gender through language. As we will see, the generalizations about word classes discussed in the previous sections hold for some clusters, but are reversed in others.

We apply probabilistic clustering in order to group authors who are linguistically similar; specifically, we represent each author as a list of word counts across the same vocabulary of 10,000 words that are used in the classification experiment. The clustering algorithm, based on the Expectation-Maximization framework (Hastie, Tibshirani, and Friedman 2009), is an iterative algorithm that groups authors together by similarities in word usage. After first randomly assigning all authors to one of twenty clusters, the algorithm then alternates between (1) calculating the center of the cluster (from the average word counts of all authors who have been assigned to it) and (2) assigning each author to the nearest cluster, based on the distance between their word counts and the average word counts that define the cluster center. This procedure is described in detail in Appendix I. The original dataset is 56% male, but in the clustering analysis we randomly subsample the male authors so that the gender proportions are equal.

The clusters can be used to identify sets of words that are often used by the same authors. While factor analysis is sometimes employed for this purpose (e.g., Argamon et al. 2007), it has two drawbacks. First, it describes each author in terms of a set of real-valued factor loadings, rather than membership in a single cluster (so that it does not identify discrete “groups” of authors). Second, standard factor analysis makes Gaussian assumptions that are inappropriate for word count data: word counts are non-negative integers, while the Gaussian distribution applies to all real numbers. In our approach, the characteristic words for each cluster can be identified by the odds ratio, which is computed by taking each word's probability among authors in the cluster, and dividing by its probability over the entire dataset.

Appendix II shows the ten most characteristic words for each cluster, omitting three clusters with fewer than 100 authors. Even though the clusters were built without considering gender, most have strong gender orientations. Of the seventeen clusters shown, fourteen skew at least 60% female or male; for even the smallest reported cluster (C19, 198 authors), the chance probability of a 60/40 gender skew is well below 1%. This shows that even a purely text-based division of authors yields groupings that are strongly related to gender. However, the cluster analysis allows for multiple expressions of gender, which may reflect interactions between gender and other categories such as age or race. For example, contrast the female-dominated clusters C14 and C5, or the male-dominated clusters C11 and C13; indeed, nearly every one of these clusters seems to tell a demographic story. This underscores the importance of intersectionality — the impossibility of pulling different dimensions of social life like gender, race, and affect into separate strands (Crenshaw 1991; McCall 2005; Brah and Phoenix 2004).

The clusters of words shown in Appendix II may suggest a distinction between topic (e.g., politics and sports) and style (e.g., emoticons, swears). But such a distinction is difficult to operationalize. Certainly there are many words which cannot be marked as purely stylistic or topical: for example, *basketball*, *bball*, and *hoops* all refer to a very specific topic, yet each also carries its own stylistic marking. More generally, Eckert (2008) casts doubt on the possibility of pulling style and topic apart, as 'different ways of saying things are intended to signal different ways of being, which includes different potential things to say.' The clusters in Appendix II group authors based on patterns of lexical co-occurrence, and it may be more useful to think of these clusters in terms of *verbal repertoires*, which involve a complex intersection of social positions (Gumperz 1964). These quantitative patterns suggest repertoires that mix identities, styles, and topics, and these collections of words and authors may be fruitful sites for deeper qualitative analysis.

Several clusters reverse the findings about the relationship between gender and various word classes summarized in Table 2. For example, we saw that women on aggregate used significantly fewer dictionary words than men, and significantly more non-dictionary words (excluding named entities). Yet the most female-dominated cluster (C14, which is 90% women) uses dictionary words at a significantly higher rate than men (75.6 to 74.9 per 100 words; the rate is 74.2 for women overall), and uses pronounceable non-dictionary words at a significantly lower rate than men (1.93 to 3.35; the rate is 3.55 for women overall). An analysis of the top words associated with this cluster suggests that its members may be older (the top word, *hubs*, is typically used as a shortening for husband). Cluster C4 (63% women) displays similar tendencies, but also uses significantly fewer abbreviations and emoticons than men: 1.84 for this mostly female cluster, 2.99 for men overall, and 4.28 for women overall.

Among the male-dominated clusters, C11 defies the aggregate gender trends on dictionary words (69.0 to 74.2 for women and 74.9 for men overall), unpronounceable non-standard terms (5.95 to 4.28 for women, 2.99 for men overall), and pronounceable non-standard terms (11.2, by far the most of any cluster). This cluster captures features of African-American English: *finna* is a transcription of *fixing to* (Rickford and Rickford 2000; Green 2002); the abbreviations *lls* and *lmaoo* have been previously shown to be more heavily used in messages from zip codes with high African-American populations (Eisenstein, Smith, and Xing 2011). Cluster C9 also features several terms that appear to be associated with African-American English, but it displays a much lower rate of taboo terms than C11, and is composed of 60% women.

Taboo terms are generally preferred by men (0.69 versus 0.47 per hundred words), but several male-associated clusters reverse this trend: C10, C13, C15, and C20 all use taboo terms at significantly lower rates than women overall. Of these clusters, C10 and C15 seems to suggest work-related messages from the technology and marketing spheres, where taboo language would be strongly inhibited. C13 and C18 are alternative sports-related clusters; C13 avoids taboo words and non-standard words in general, while C18 uses both at higher rates, and includes mentions of the hiphop performers *@macmiller*, *@wale* and *@fucktyler*.

All of the male-associated clusters mention named entities at a higher rate than women overall, and all of the female-associated clusters mention them at a lower rate than men overall. The highest rate of named entities is found in C13, an 89% male cluster whose top words are almost exclusively composed of athletes and sports-related organizations. Similarly, C20 (72.5% male) focuses on politics, and C15 focuses on technology and marketing-related entities. While these clusters are skewed towards male authors, they contain sizable minorities of women, and these women mention named entities at a rate comparable to the cluster as a whole — well above the average rate for men overall. The tightly-focused subject matter of the words characterizing each of these clusters suggests that the topic of discourse plays a crucial role in mediating between gender and the frequent mention of named entities, just as Herring and Paolillo (2006) found genre to play a similar mediating role in blogs. In our data, men seem more likely to communicate about hobbies and careers that relate to large numbers of named entities, and this, rather than a generalized preference for ‘informativity’ or ‘explicitness’, seems the most probable explanation for the demonstrated male tendency to mention named entities more often than women in our data.

Social categories such as gender cannot be separated from other aspects of identity (see also Bourdieu 1977; Giddens 1984; Sewell, Jr. 1992). For example, while technology and sports clusters both skew disproportionately male, it seems unlikely that masculinity has the same meaning in each domain. Cluster analysis reveals the danger in seemingly innocuous correlations between linguistic resources and high-level social categories such as gender. If we start with the assumption that 'female' and 'male' are the relevant categories, then our analyses are incapable of revealing violations of this assumption. Such an approach may be adequate if the goal is simply to predict gender based on text. However, when we turn to a descriptive account of the interaction between language and gender, this analysis becomes a house of mirrors, which by design can only find evidence to support the underlying assumption of a binary gender opposition.

While most of the clusters are strongly gendered, none are 100% male or female. What can we say about the 1,242 men who are part of female-dominated clusters and the 1,052 women who are part of male-dominated clusters? These individuals could be dismissed as outliers or statistical noise. Because their language aligns more closely with the other gender, they are particularly challenging cases for machine learning. But rather than ask how we can improve our algorithms to divine the 'true' gender of these so-called outliers, we might step back and ask how their linguistic choices participate in the construction of gendered identities. The cluster analysis suggest habitual stances that we might label as 'mother', 'bff', 'politico', or 'sports fanatic'. What we understand as gender is built up indirectly, with many ways to perform 'male' or 'female'. As we will see in the next section, far from being statistical noise, the language patterns of 'outlier' individuals fit coherently into a larger picture of online social behavior.

## GENDER HOMOPHILY IN ONLINE SOCIAL NETWORKS

Corpus statistics like word counts are built out of situated, context-rich individual uses. Stances constantly shift as we talk to different people, about different things, and call up different selves to do the talking (Du Bois 2007). A full reckoning of the implications of the stancetaking model for corpus linguistics is beyond the scope of this work, but clearly some consideration of the audience is needed if we are to understand how language expresses social variables such as gender. As a first step in this direction, we compare the use of gendered language with the aggregate gender composition of the social networks of the individuals in our corpus.

The theory of homophily — 'birds of a feather flock together' — has been demonstrated to have broad applicability across a range of social phenomena (McPherson, Smith-Lovin, and Cook 2001). This theory applies to social media, where it is possible to make accurate predictions of a range of personal attributes based on the attributes of nearby individuals in the social network (e.g., Thelwall 2008). The social network in our Twitter data displays significant gender homophily: 63% of mutual-@ connections are between same-gender individuals.

Thus, gender is correlated both with linguistic resources as well as with social network composition. On the view that language and social network connections depend only on the author gender, we would expect these two channels to be *conditionally independent* given gender. In contrast, a more multifaceted model of gender would not be committed to viewing language and social behavior as conditionally independent. Sociolinguistic work on the relationship between language and social networks finds that individuals with stronger ties to their local geographical region make greater use of local speech variables (Bortoni-Ricardo 1985; Gal 1979; Milroy 1991). We ask whether a similar phenomenon applies to gender: do individuals with a greater proportion of same-gender ties make greater use of gender-marked variables in social media?

We construct an undirected social network from direct conversations in our data; details are found earlier in the paper. As above, we identify an individual's local network as skewed using a statistical hypothesis test that compares how strongly the gender composition of each individual's network departs from 50% (representing the null hypothesis of no homophily, an even balance of women and men). Specifically, we measure the cumulative distribution function (CDF) of the observed counts of men and women under a binomial distribution with $p = .50$ and $n$ = the number of friends. This lets us find, for example, that under the null hypothesis of homophily, the probability that a set of 25 people would have 4 or fewer men is 0.0005.

[Insert Table 3 about here]

We find a strong correlation between the use of gendered language and the gender skew of social networks. Our text-based gender classifier quantifies the extent to which each author’s language use coheres with the aggregated statistics for men and women. Table 3 presents the Pearson correlation between the gender composition of an individual's social network and the probability output by the text-based gender classifier. This correlation is statistically significant for both women and men (*r*=.38 and *r*=.33, respectively). Similar correlations obtain between the gender composition of an individual's network and the use of gendered lexical markers (*r*=.34 for women, *r*=.45 for men). The more gendered an author's language (in terms of aggregated statistics), the more gendered the social network.

[Insert Figure 1 about here]

Figure 1 presents this information graphically, by illustrating the relationship between the confidence of our gender classifier and the social network gender distribution of the individuals whose gender we predict. Here we group the classifier confidence into ten bins and plot the gender network composition for the authors in each bin. On average, the women in our dataset have social networks that are 58% female. However, for the decile of women whose language is most strongly marked as female by the classifier, the average network composition is 77% female. The decile of women whose language is least strongly marked as female have networks that are on average 40% female. Similarly, the average male in our dataset has a social network that is 67% male, but in the extreme deciles, the average social networks are 78% and 49% male respectively. Besides the classifier, we obtain similar findings with the 1,000 lexical gender markers: the usage of same-gender markers increases with the proportion of same-gender friends. Overall, these results paint a consistent picture, in which the use of gendered language resources parallels the gender composition of the social network.

[Insert Figure 2 about here]

Finally, we ask whether the gender composition of an author's social network offers any new information about gender, beyond the information carried by language. To measure this, we add features about the social network composition to the text-based gender classifier. Figure 2 shows the results; classifier accuracy is on the x-axis, and the y-axis shows the effect of varying the maximum number of word tokens per author; as above, cross-validation is used to select the regularization term (see Appendix I). In the limit of no text, the accuracy with network features is 63% (corresponding to the degree of gender homophily), and the accuracy without network features or text is 56% (corresponding to the total proportion of male authors in the dataset). Network features are informative when a very limited amount of text is available, but their impact soon disappears: given just 1,000 words per author, network features no longer offer any observable improvement in accuracy, even though the classifier has not reached a ceiling and can still be improved by adding text.

Ambiguity in the use of linguistic resources is not statistical noise, but rather the signature of individuals who have adopted stances and personae at odds with mainstream gender norms. These stances and personae shape social network connections just as they shape the use of linguistic resources. We find theoretical support for this analysis in both accommodation and audience design (Bell 1984; Clark 1996; Giles and Coupland 1991), which suggest that individuals will often modulate their language patterns to match those of their interlocutors. On this view, language does not *reveal* gender as a binary category; rather, linguistic resources are used to position oneself relative to one's audience. Gender emerges indirectly, and to the extent that a linguistic resource indexes gender, it is pointing to (and creating) the habitual, repeated, multifaceted positionings inherent in every situated use of language.

## DISCUSSION

We began with an approach that has become widespread in computational social media analysis: the presentation of a highly accurate predictive model, followed by a post hoc analysis of the power of various linguistic features. In the case of gender, it is tempting to assemble results about lexical frequencies into larger narratives about very broad stylistic descriptors, such as a gendered preference for language that conveys 'involvement' or 'information'. By building our model from individual word counts, we avoid defining these broad descriptors from prior assumptions about the high-level pragmatic function of words or word classes; instead we let the data drive the analysis. But the same logic that leads us to question the identification of, say, all nouns as 'informational' also leads us to revise our analysis of the social variable. A quantitative approach built around a binary gender opposition can only yield results that support and reproduce this underlying assumption. While the statistical relationships between word frequencies and gender categories are real, they are but one corner of a much larger space of possible results that might have been obtained had we started with a different set of assumptions.

Gender is a powerful force in structuring our social lives, and one cannot deny the social reality of 'male' and 'female' social categories. But categories are never simply descriptive; they are normative statements that draw lines around who is included and excluded (Butler 1990). Computational and quantitative models have often treated gender as a stable binary opposition, and in so doing, have perpetuated a discourse that treasures differences over similarities, and reinforces the ideology of the status quo. It is not a theoretical innovation to suggest that gender is more complicated than two categories. What our analysis adds is a demonstration of how such models can be descriptively inadequate.

While richer treatments of gender have typically been supported by qualitative rather than quantitative analysis, we see the convergence of machine learning and large social datasets as offering exciting new opportunities to investigate how gender is constructed, and how this construction is manifested in different contexts. Machine learning offers a bountiful harvest of modeling techniques that minimize the need for categorical assumptions. These models permit exploratory analysis that reveals patterns and associations that might have been rendered invisible by less flexible hypothesis-driven analysis. We are especially interested in quantitative models of how social variables like gender are constructed and reproduced in large numbers of individual interactions. In this paper, cluster analysis has demonstrated the existence of multiple gendered styles, stances, and personae; we hope that a more nuanced model might allow statistical reasoning on the level of individual micro-interactions, thus yielding new insights about the various settings and contexts in which gender is manifested in and constructed by language.

# APPENDIX I: COMPUTATIONAL AND QUANTITATIVE METHODS

**Identifying gender markers.** Our goal is to identify words that are used with unusual frequency by authors of a single gender. Assume that each term has an unknown likelihood $f_i$, indicating the proportion of authors who use term $i$. For gender $j$, there are $N_j$ authors, of whom $k_{ji}$ use term $i$; the total count of the term $i$ is $k_i$. We ask whether the count $k_{ji}$ is significantly larger than expected. Assuming a non-informative prior distribution on $f_i$, the posterior distribution (conditioned on the observations $k_i$ and $N$) is Beta($k_i$, $N$-$k_i$). The distribution of the gender-specific counts can be described by an integral over all possible $f_i$. This integral defines the Beta-Binomial distribution (Gelman, Carlin, Stern, and Rubin 2004), and has a closed form solution. We mark a term as having a significant gender association if the cumulative distribution at the count $k_{ji}$ is $p < .05$.

**Clustering**. We find clusters of authors using the expectation-maximization (EM) algorithm (Hastie et al. 2009). Each author is assigned a distribution over clusters; each cluster has a probability distribution over word counts and a prior strength. In the EM algorithm, these parameters are iteratively updated until convergence. The probability distribution over words uses the Sparse Additive Generative Model (Eisenstein, Ahmed, and Xing 2011), which is especially well suited to high-dimensional data like text. For simplicity, we perform a hard clustering, sometimes known as hard EM. Since the EM algorithm can find only a local optimum, we make 25 runs with randomly-generated initial assignments, and select the run with the highest likelihood.

Public software implementations of these computational methods can be found at

https://github.com/jacobeisenstein/jos-gender-2014.

[i] Acknowledgments: The research reported in this article was support by an ARCS Foundation scholarship to D.B. and NSF grant IIS-1111142 to J.E. We thank the anonymous reviewers for their constructive comments and we'd also like to thank Alexandra D'Arcy, Natalia Cecire, Penny Eckert, Carmen Fought, Scott Kiesling, Kyuwon Moon, Brendan O'Connor, John Rickford, and Noah A. Smith for helpful discussions. This work was made possible through the use of computing resources made available by the Open Cloud Consortium and Yahoo.

**Appendix II**: Clusters based only on words-in-common, sorted by % of female authors. Highlighting demonstrates clusters that are highly gendered-skewed but whose patterns reverse the trends obtained from gender-only classification.

| | **Size** | **% fem** | **Dict** | **Punc** | **UnPron** | **Pron** | **NE** | **Num** | **Taboo** | **Hash** | **Top words** |
|---|---|---|---|---|---|---|---|---|---|---|---|
| Skews… | | | M | F | F | F | M | M | M | M | |
| c14 | 1,345 | 89.60% | 75.58% | 16.44% | 3.27% | 1.93% | 1.66% | 0.85% | 0.14% | 0.13% | hubs blogged bloggers giveaway @klout recipe fabric recipes blogging tweetup |
| c7 | 884 | 80.40% | 73.99% | 13.13% | 5.27% | 4.27% | 1.99% | 0.83% | 0.37% | 0.16% | kidd hubs xo =] xoxoxo muah xoxo darren scotty ttyl |
| c6 | 661 | 80.00% | 75.79% | 16.35% | 3.07% | 2.15% | 1.54% | 0.70% | 0.32% | 0.09% | authors pokemon hubs xd author arc xxx ^_^ bloggers d: |
| c16 | 200 | 78.00% | 70.98% | 14.98% | 6.97% | 3.45% | 2.19% | 0.90% | 0.10% | 0.43% | xo blessings -) xoxoxo #music #love #socialmedia slash :)) xoxo |
| c8 | 318 | 72.30% | 73.08% | 9.09% | 7.30% | 7.06% | 1.96% | 0.80% | 0.56% | 0.15% | xxx :') xx tyga youu (: wbu thankyou heyy knoww |
| c5 | 539 | 71.10% | 71.55% | 14.64% | 5.84% | 4.29% | 1.94% | 0.82% | 0.77% | 0.16% | (: :') xd (; /: <333 d: <33 </3 -___- |
| c4 | 1,376 | 63.00% | 77.09% | 15.81% | 1.84% | 1.82% | 2.02% | 0.78% | 0.52% | 0.12% | && hipster #idol #photo #lessambitiousmovies hipsters #americanidol #oscars totes #goldenglobes |
| c9 | 458 | 60.00% | 70.48% | 10.49% | 7.49% | 7.70% | 2.00% | 0.89% | 0.65% | 0.30% | wyd #oomf lmbo shyt bruh cuzzo #nowfollowing lls niggas finna |
| c19 | 198 | 58.10% | 70.25% | 21.77% | 3.72% | 2.24% | 1.28% | 0.31% | 0.36% | 0.07% | nods softly sighs smiles finn laughs // shrugs giggles kisses |
| c17 | 659 | 55.80% | 72.30% | 12.84% | 4.78% | 5.62% | 1.82% | 0.65% | 1.69% | 0.30% | lmfaoo niggas ctfu lmfaooo wyd lmaoo nigga #oomf lmaooo lmfaoooo |
| c1 | 739 | 46.00% | 75.38% | 16.31% | 3.15% | 1.60% | 2.25% | 1.02% | 0.11% | 0.18% | qr /cc #socialmedia linkedin #photo seo webinar infographic klout |
| c15 | 963 | 34.70% | 74.62% | 15.40% | 3.29% | 2.42% | 2.74% | 1.05% | 0.32% | 0.17% | #photo /cc #fb (@ brewing #sxsw @getglue startup brewery @foursquare |
| c20 | 429 | 27.50% | 75.38% | 16.74% | 2.09% | 1.41% | 3.10% | 0.91% | 0.23% | 0.14% | gop dems senate unions conservative democrats liberal palin republican republicans |
| c11 | 432 | 26.20% | 68.97% | 8.32% | 5.95% | 11.16% | 2.01% | 0.88% | 2.32% | 0.38% | niggas wyd nigga finna shyt lls ctfu #oomf lmaoo lmaooo |
| c18 | 623 | 18.90% | 77.46% | 10.47% | 2.75% | 4.40% | 2.84% | 1.07% | 0.82% | 0.19% | @macmiller niggas flyers cena bosh pacers @wale bruh melo @fucktyler |
| c10 | 1,865 | 14.60% | 77.72% | 16.17% | 1.51% | 1.27% | 2.03% | 0.89% | 0.34% | 0.06% | /cc api ios ui portal developer e3 apple's plugin developers |
| c13 | 761 | 10.60% | 75.92% | 15.12% | 1.60% | 1.67% | 3.78% | 1.44% | 0.36% | 0.10% | #nhl #bruins #mlb nhl #knicks qb @darrenrovell inning boozer jimmer |

Table 1: Comparison of gender markers with previous research; 'ns' indicates no significant association; 'mixed' indicates markers for male and female genders.

| | Previous literature | Our analysis |
|---|---|---|
| Pronouns | F | F |
| Emotion terms | F | F |
| Kinship terms | F | Mixed |
| CMC words (*lol*, *omg*) | F | F |
| Conjunctions | F | ns |
| Clitics | F | ns |
| Articles | M | ns |
| Numbers | M | M |
| Quantifiers | M | ns |
| Technology words | M | M |
| Prepositions | Mixed | ns |
| Swear words | Mixed | M |
| Assent | Mixed | F |
| Negation | Mixed | Mixed |
| Emoticons | Mixed | F |
| Hesitation | Mixed | F |

Table 2: Word category frequency by gender. All differences statistically significant at $p < .01$.

| | **F** | **M** |
|---|---|---|
| Standard dictionary | 74.20% | 74.90% |
| Punctuation | 14.60% | 14.20% |
| Non-standard, not pronounceable (e.g., *:), lmao*) | 4.28% | 2.99% |
| Non-standard, pronounceable (e.g., *luv)* | 3.55% | 3.35% |
| Named entities | 1.94% | 2.51% |
| Numbers | 0.83% | 0.99% |
| Taboo | 0.47% | 0.69% |
| Hashtags | 0.16% | 0.18% |

Table 3: Pearson correlations between the gender composition of author social networks and the use of gendered language, as measured by classifier confidence and the proportion of gendered markers; the more gendered an author's language, the more gendered the social network. Confidence intervals are 99%.

| | Female authors | Male authors |
|---|---|---|
| Classifier vs. network composition | 0.38 ($0.35 \leq r \leq 0.40$) | 0.33 ($0.3 \leq r \leq 0.36$) |
| Markers vs. network composition | 0.34 ($0.31 \leq r \leq 0.37$) | 0.45 ($0.43 \leq r \leq 0.47$) |

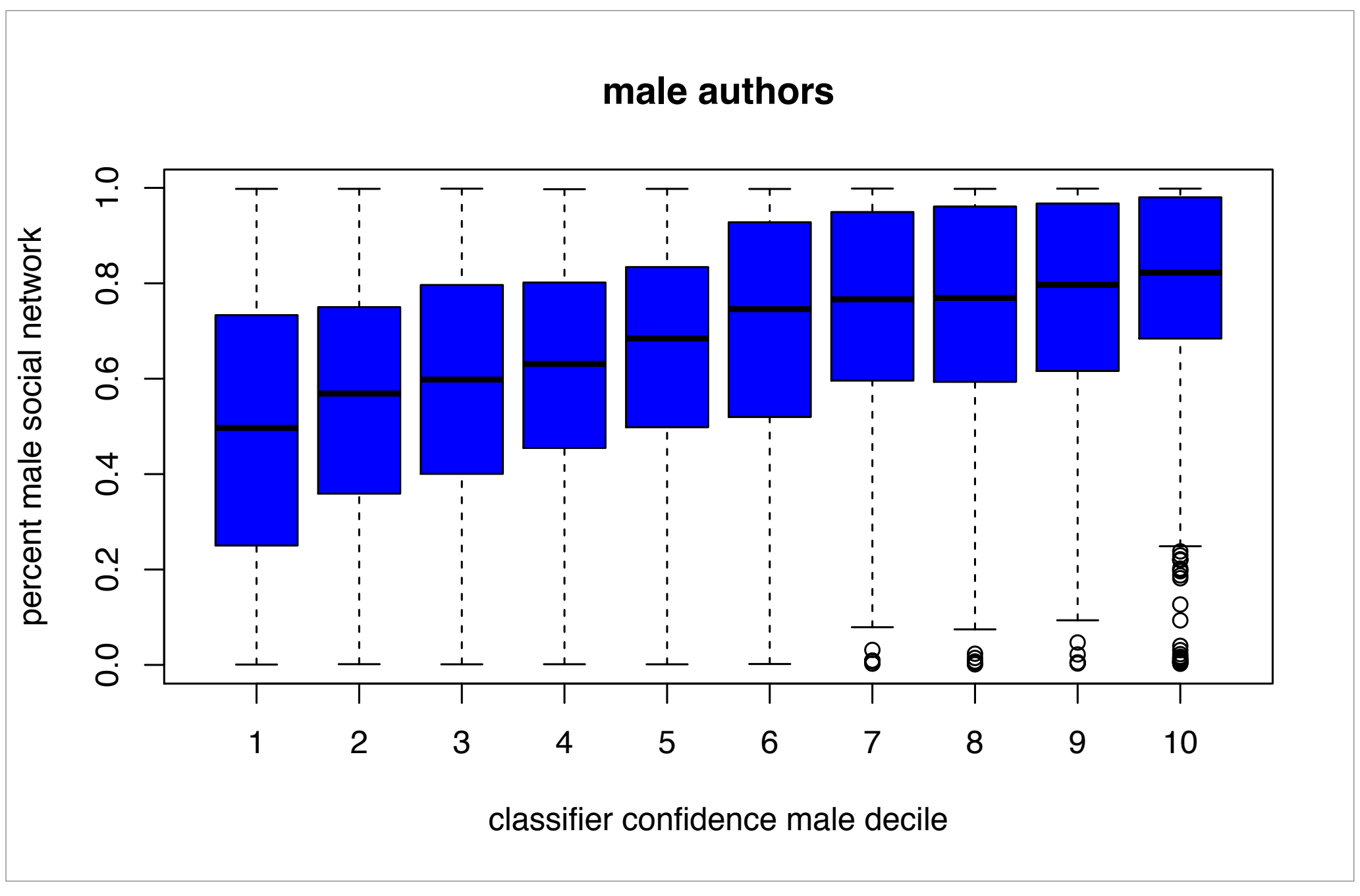

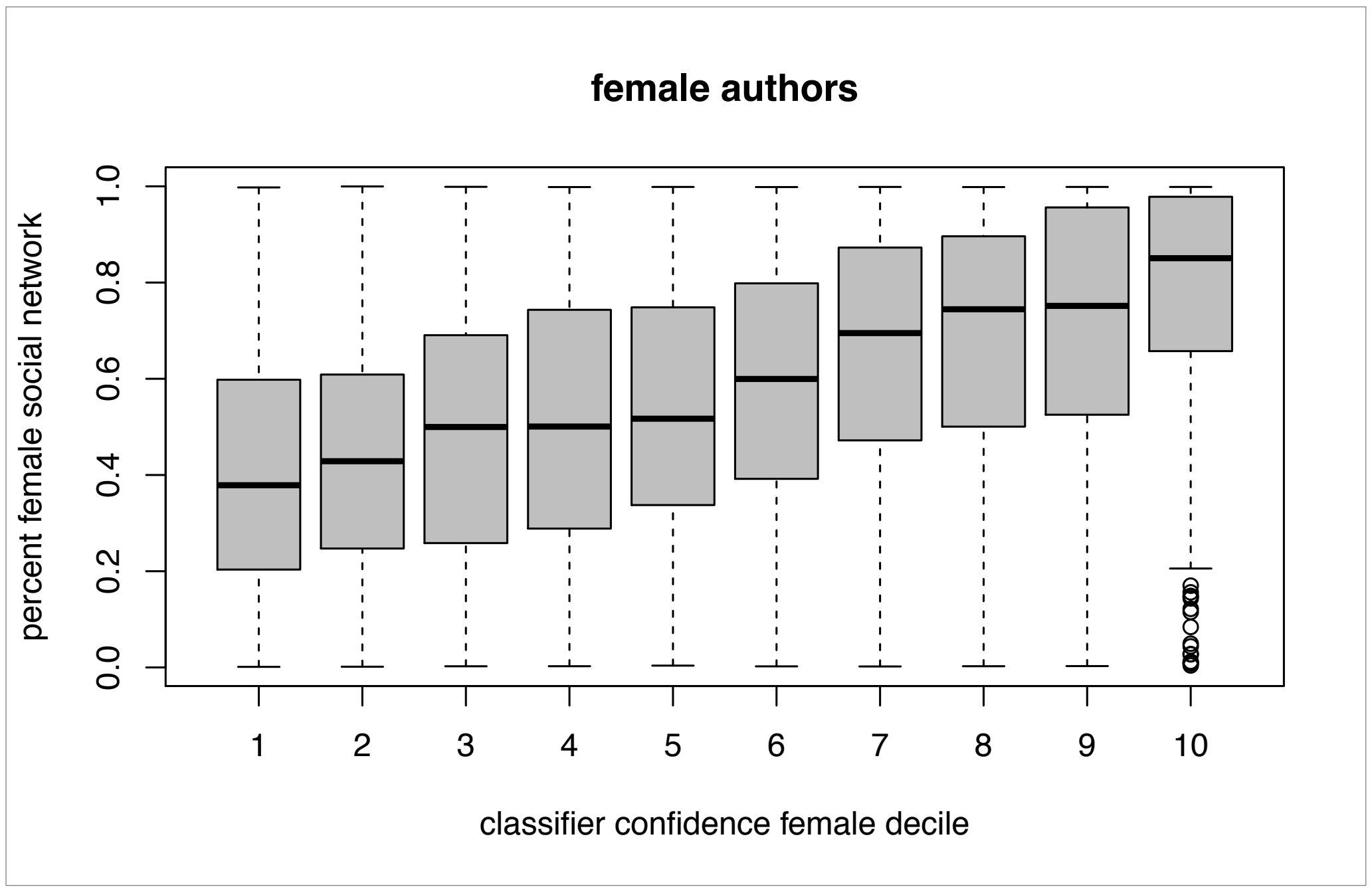

Figure 1. Social network composition and gender classifier confidence, binned by decile; the higher the gender skew of the social network, the more confident the classifier is in its prediction.

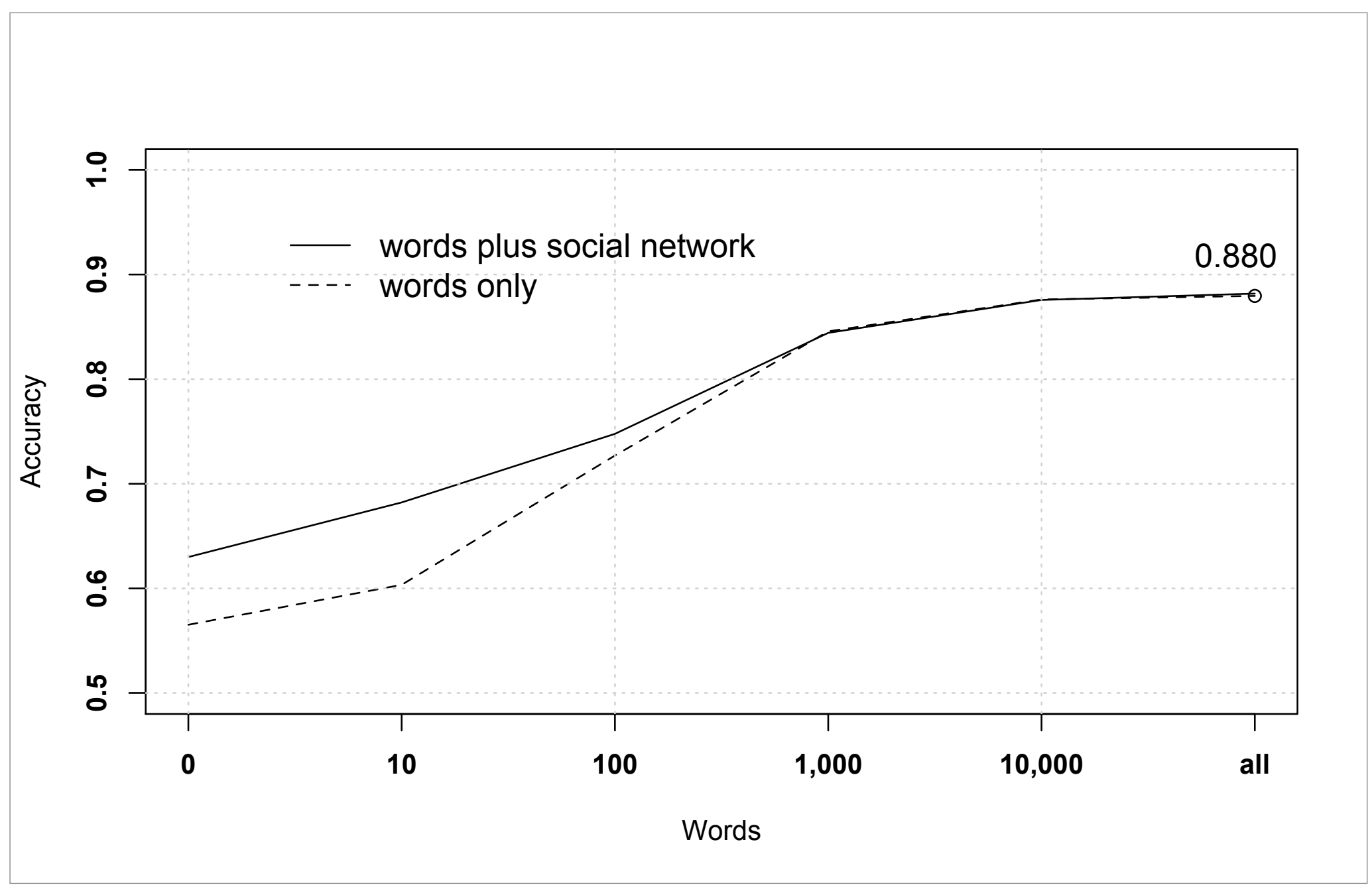

Figure 2. Gender prediction accuracy, plotted against the number of words seen per author. Social network information helps when there is little text, but in the limit it adds no new information.

**Address correspondence to:**

Tyler Schnoebelen

Idibon, Inc.

870 Market Street, Suite 828

San Francisco, CA 94102

tyler@idibon.com